\documentclass[runningheads]{llncs}
\usepackage{graphicx}
\usepackage{amsmath,amssymb} 
\usepackage{color}
\usepackage[width=122mm,left=12mm,paperwidth=146mm,height=193mm,top=12mm,paperheight=217mm]{geometry}

\usepackage{times}
\usepackage{epsfig}
\usepackage{amsmath}
\usepackage{algorithm}
\usepackage{algorithmic}
\usepackage{amssymb}
\usepackage{multirow}
\usepackage[bold]{hhtensor}
\usepackage{subfig}
\newcommand{\squishlist}{
 \begin{list}{$\bullet$}
  { \setlength{\itemsep}{0pt}
     \setlength{\parsep}{1pt}
     \setlength{\topsep}{1pt}
     \setlength{\partopsep}{0pt}
     \setlength{\leftmargin}{1.5em}
     \setlength{\labelwidth}{1em}
     \setlength{\labelsep}{0.5em} } }

\newcommand{\squishend}{
  \end{list}  }
  \newtheorem{assumption}{Assumption}
\usepackage[pagebackref=true,breaklinks=true,letterpaper=true,colorlinks,bookmarks=false]{hyperref}

\begin{document}
\pagestyle{headings}
\mainmatter

\title{Improving Multi-label Learning with Missing Labels by\\ Structured Semantic Correlations} 

\titlerunning{Structured Semantic Correlations}

\authorrunning{Yang, Zhou and Cai}

\author{Hao Yang \inst{1} \and Joey Tianyi Zhou \inst{2} \and Jianfei Cai \inst{3}}


\institute{Rolls-Royce@NTU Corp Lab, Singapore. \email{lancelot365@gmail.com} 
\and 
IHPC, A*STAR, Singapore. \email{zhouty@ihpc.a-star.edu.sg} \\
\and
School of Computer Science and Engineering, NTU, Singapore. \email{ASJFCai@ntu.edu.sg}\\}

\maketitle

\begin{abstract}
Multi-label learning has attracted significant interests in computer vision recently, finding applications in many vision tasks such as multiple object recognition and automatic image annotation. Associating multiple labels to a complex image is very difficult, not only due to the intricacy of describing the image, but also because of the incompleteness nature of the observed labels. Existing works on the problem either ignore the label-label and instance-instance correlations or just assume these correlations are linear and unstructured. Considering that semantic correlations between images are actually structured, in this paper we propose to incorporate structured semantic correlations to solve the missing label problem of multi-label learning. Specifically, we project images to the semantic space with an effective semantic descriptor. A semantic graph is then constructed on these images to capture the structured correlations between them. We utilize the semantic graph Laplacian as a smooth term in the multi-label learning formulation to incorporate the structured semantic correlations. Experimental results demonstrate the effectiveness of the proposed semantic descriptor and the usefulness of incorporating the structured semantic correlations. We achieve better results than state-of-the-art multi-label learning methods on four benchmark datasets.
\end{abstract}

\section{Introduction}
\label{intro} 
\begin{figure}
\centering \captionsetup[subfigure]{labelformat=empty}
  \subfloat[\textbf{animal}, \textbf{clouds}, \textit{grass}, \textit{green}, \textbf{lake}, \textit{landscape}, \textbf{plantlife}, \textit{reindeer}, \textbf{sky}, \textbf{water}]{\includegraphics[width=0.35\textwidth]{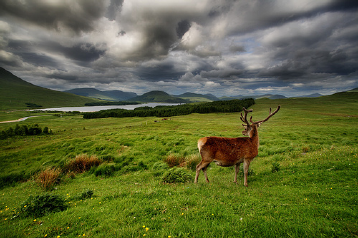}} \qquad
  \subfloat[\textit{Audi}, \textbf{car}, \textbf{structures}, \textit{racing}, \textit{road}, \textit{track}, \textbf{transport}]{\includegraphics[width=0.35\textwidth]{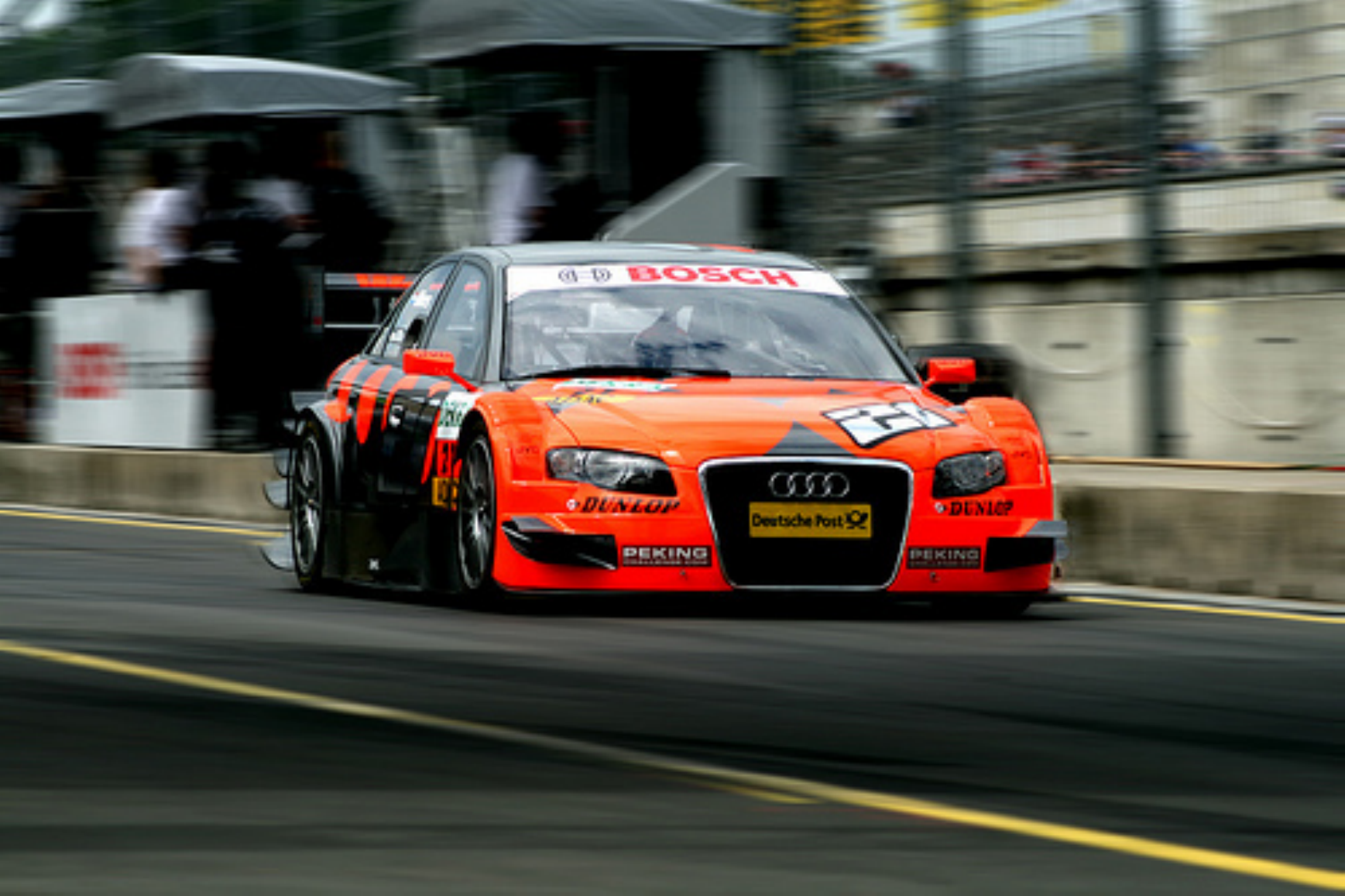}}
  \caption{Example labels from \textsc{Flickr25K} dataset. The bold face labels are original annotations from the users. The italic labels are other possible labels. These examples illustrate the missing labels problem of multi-label learning.}
  \label{example-incomplete-labels}
\end{figure}

Multi-label learning has been an important research topic in machine learning~\cite{Bi2014,Liu2015,Yu2014} and data mining~\cite{Kong2013,Kong2014}. Unlike conventional classification problems, in multi-label learning each instance can be associated with multiple labels simultaneously. During recent years, multi-label learning has been applied on many computer vision tasks, especially on visual object recognition~\cite{VOC,Gong2013,OquabCVPR2014} and automatic image annotation~\cite{Cabral2011,Tariq2015,Wang2014}. In addition to the difficulty of assigning multiple labels/tags to complex images, multi-label learning often encounters the problem of incomplete labels. In real world scenarios, since the number of possible labels/tags is often very large (could be as large as the whole vocabulary set) and there often exist ambiguities among labels (e.g, ``car'' vs ``SUV''), it is very difficult to obtain a perfectly labeled training set. Fig.~\ref{example-incomplete-labels} shows some examples of annotations from \textsc{Flickr25K} dataset. We can see that many possible labels are missing as it is impossible for labelers to go through the entire vocabulary set to extract all proper tags.

Due to the incompleteness nature of multi-label learning, many methods have been proposed to solve the problem of multi-label learning with missing labels. Most existing works focus on exploiting the correlations between features and labels (feature-label correlations)~\cite{Tsoumakas2007}, the correlations between labels (label-label correlations) and the correlations between instances (instance-instance correlations)~\cite{Bi2014,Yu2014,Chen2013,Cabral2011}. Binary relevance (BR) ~\cite{Tsoumakas2007} is a popular baseline for multi-label classification, which simply treats each class as a separate binary classification and makes use of feature-label correlations to solve the problem. However, its performance can be subpar as it ignores the correlations between labels and between instances. Several matrix completion based methods~\cite{Xu2013,Kong2014,Yu2014} handle the missing labels problem by implicitly exploiting label label correlations and instance-instance correlations with low-rank regularization on the label matrix. FastTag~\cite{Chen2013} also implicitly utilizes label-label correlations by learning an extra linear transformation on the label matrix to recover possible missing labels. On the other hand, LCML~\cite{Bi2014} explicitly handles missing labels with a probabilistic model.

Although these existing works exploit the correlations for learning classifiers and recovering missing labels, they generally (implicitly) assume that those correlations are linear and unstructured. However, in real world applications, especially image recognition, the label-label correlations and instance-instance correlations are actually structured. For example, label ``landscape'' is likely to co-exist with labels like ``sky'', ``mountain'', ``river'', etc, but it is not likely to co-exist with ``desk'', ``computer'', ``office'', etc. Deng et al.~\cite{Deng2014} already shows that the structured \emph{label} correlations can benefit \emph{multi-class classification}. In this work, we focus on exploiting the structured correlations between \emph{instances} to improve \emph{multi-label learning}. Given proper prior knowledge, our framework can also incorporate structured label correlations easily.

The key to utilize structured instance-instance correlations is to make use of semantic correlations between images, as \emph{semantically similar images should share similar labels}. If we can effectively extract good semantic representations from images, we should be able to capture the structured correlations between instances.

A semantic representation of an image is a high level description of the image. One popular semantic representation is based on the score vectors of the classifier outputs. Many works have discussed the potential of such representations~\cite{Bergamo2014,Dixit2015,Kwitt2012,Lampert2014,Su2012}. For example, Su and Jurie~\cite{Su2012} proposed to use bag of semantics (BoS) to improve the image classification accuracy. Lampert et al.~\cite{Lampert2014} employed semantics representations to describe objects by their attributes. Dixit et al.~\cite{Dixit2015} combined CNN (\emph{convolutional neural networks}) activations, semantic representations and Fisher vectors to improve scene classification. Kwitt et a.~\cite{Kwitt2012} also proposed to apply semantic representations on manifold for scene classification.

In this paper, we propose a new semantic representation, which is the concatenation of a global semantic descriptor and a local semantic descriptor. The global part of our semantic representation is similar to~\cite{Dixit2015}, which is the object-class posterior probability vector extracted from CNN trained with \textsc{ILSVRC 2012} dataset. The global semantic descriptor describes ``what is the image in general'' according to a large number of concepts developed in the general large-scale dataset. We also introduce a local semantic descriptor extracted by averagely pooling the labels/tags of visual neighbors of each image in the specific target domain. The local semantic descriptor describes ``what does the image specifically look like''. By combining the global and the local semantic descriptors, we achieve more accurate semantic representation.

With the accurate semantic descriptions of images, we propose to incorporate semantic instance-instance correlations to the multi-label learning problem by adding structures via graph. To be specific, after projecting the images into semantic space, we consider each semantic representation as a node and the whole image set as an undirected graph. Each edge of the graph connects two semantic image representations, and its weight represents the similarity between the node pair. We introduce the semantic graph Laplacian as a smooth term in the multi-label learning formulation to incorporate structured instance-instance correlations captured by the semantic graph. Experiments on four benchmark datasets
demonstrate that by incorporating structured instance-instance semantic correlations, our proposed method significantly outperforms the state-of-the-art multi-label learning methods, especially at low observed rates of training labels (e.g. only observing $10\%$ of the given training labels). The major contributions of this paper lie in the proposed semantic representation and the proposed method to incorporate structured semantic correlations into multi-label learning.

\section{Related Works on Multi-label Learning}
\label{related} 
Binary Relevance (BR)~\cite{Tsoumakas2007} is a standard baseline for multi-label learning, which treats each label as an independent binary classification. Linear or kernel classification tools such as \textsc{LIBLINEAR}~\cite{Fan2008} can then be applied to solve each binary classification subproblem. Although in general BR can achieve certain accuracy for multi-label learning tasks, it has two drawbacks. First of all, BR ignores the correlations between labels and between instances, which could be helpful for recognition. Secondly, as the label set size grows, the computational cost for BR in both training and testing becomes infeasible. To solve the first problem, some researchers proposed to estimate the label correlations from the training data. In particular, Hariharan et al.~\cite{Hariharan2010} and  Petterson and Caetano~\cite{Petterson2011} represent label dependencies by pairwise correlations computed from the training set, but such representations could be crude and inaccurate if the distribution of the training data is biased. LCML~\cite{Bi2014} uses a probability model to explicitly handle the label correlations. In multi-class classification,~\cite{Deng2014} exploits external label relation graph to model the correlations between labels. There also exist some works~\cite{Kong2013,Kong2014,Xu2013} that use the idea of matrix completion to implicitly deal with label correlations by imposing a nuclear norm to the formulation. To solve the second problem of BR, PLST~\cite{Tai2012} and CPLST~\cite{Chen2012} reduce the dimension of the label set by PCA related methods. Hsu et al.~\cite{Hsu2009} employs a compressed sensing based approach to reduct the label set size. In addition to reducing label set size, these methods also decorrelate the labels, thus solving the first problem to a certain degree. 

Nearest neighbors (NN) related methods are also commonly utilized in multi-label related applications. For label propagation, Kang et al.~\cite{Kang2006} proposed the Correlated label propagation (CLP) framework that propagates multiple labels jointly based on kNN methods. Yang et al.~\cite{Yang2016} utilized NN relationships as the label view in a multi-view multi-instance framework for multi-label object recognition. TagProp~\cite{Guillaumin2009} combines metric learning and kNN to propagate labels. For tag refinement, Zhu et al.~\cite{Zhu2010} proposed to use low-rank matrix completion formula with several graph constraints as the objective function to refine noisy or incomplete labels. For tag ranking, several methods~\cite{Liu2009,Jeong2013,Zhuang2011} have been proposed to learn a ranking function utilizing the correlations between tags. 

\section{Problem Formulation}
\label{formulation} In the context of multi-label learning, let
matrix $Y \in \mathbb{R}^{n\times c}$ refer to the true label
(tag) matrix with rank $r$, where $n$ is the number of instances
and $c$ is the size of label set. As $Y$ is generally not
full-rank, without loosing generality, we can assume $n \geq c
\geq r$ and $Y_{i,j} \in \left\{0,1\right\}$. Given the data set
$X \in \mathbb{R}^{n\times d}$, $n \geq d$, where $d$ is the
feature dimension of an instance. We make the following
assumption:
\begin{assumption}
\label{as:1} The column vectors in $Y$ lie in the subspace spanned
by the column vectors in $X$.
\end{assumption}
Assumption~\ref{as:1} essentially means the label matrix $Y$ can
be accurately predicted by the linear combinations of the features
of data set $X$, which is the assumption generally used in linear classification~\cite{Fan2008,Xu2013,Yu2014}.
Therefore, the goal of multi-label learning is to learn the linear
projection $M \in \mathbb{R}^{d\times c}$ such that it minimizes
the reconstruction error:
\begin{equation}
\label{regression-obj}
\begin{matrix}
\underset{M}{\min} & \left\|XM-Y\right\|_F^2,
\end{matrix}
\end{equation}
where $\left\|\cdot\right\|_{F}$ is the Frobenius norm.

Since the label matrix is generally incomplete in the real world
applications, we assume $\tilde{Y} \in \mathbb{R}^{n\times c}$ to
be the observed label matrix, where many entries are unknown. Let
$\Omega \subseteq \{1,\dots,n\}\times\{1,\dots,c\}$ denote the set
of the indices of the observed entries in $Y$, we can define a
linear operator $\mathcal{R}_{\Omega}(Y):\mathbb{R}^{n\times
c}\mapsto\mathbb{R}^{n\times c}$ as
\begin{equation}
 \tilde{Y}_{i,j} = [\mathcal{R}_{\Omega}(Y)]_{i,j}=\left\{\begin{matrix}
Y_{i,j} & (i,j)\in\Omega\\
 0& (i,j)\notin \Omega
\end{matrix}\right.
\end{equation} Then, the multi-label learning problem becomes:
given $\tilde{Y}$ and $X$, how to find the optimal $M$ so that the
estimated label matrix $XM$ can be as close to the ground-truth
label matrix $Y$ as possible.

Similar to~\cite{Xu2013,Yu2014}, we can make use of the low-rank
property of $Y$ and optimize the following objective function:
\begin{equation}
\label{completion-obj}
\begin{matrix}
\underset{M}{\min} & \lambda\|XM\|_{*} + \frac{1}{2}
\|\mathcal{R}_{\Omega}(XM)-\tilde{Y}\|_F^2,
\end{matrix}
\end{equation}
where $\left\|\cdot\right\|_{*}$ is the nuclear norm and $\lambda$
is the tradeoff parameter. \eqref{completion-obj} is
quintessentially the same as the matrix completion problem
in~\cite{Candes2009}.

Minimizing $\|XM\|_{*}$ could be intractable for large-scale
problems. If we assume that $X$ is orthogonal, which can be easily fulfilled by applying PCA to
the original data set $X$ if it is not already orthogonal, we can reformulate \eqref{completion-obj} to
\begin{equation}
\label{final-completion-obj}
\begin{matrix}
\underset{M}{\min} & \lambda\|M\|_{*} + \frac{1}{2}
\|\mathcal{R}_{\Omega}(XM)-\tilde{Y}\|_F^2
\end{matrix}
\end{equation}
so that the problem can be solved much more efficiently~\cite{Xu2013}.

The problem with~\eqref{final-completion-obj} is that by employing
the low rank condition, it implicitly assumes that rows/columns of
label matrix $Y$ is linearly dependent, i.e., the instance-instance
correlations and label-label correlations are linear and
unstructured. However, in real world applications, these
correlations are actually structured. For example, \cite{Deng2014}
has already demonstrated that structured label-label correlations
can benefit multi-class classification. In this work, we mainly
consider the structured correlations among instances, but our
framework can easily incorporate label-label correlations, if
proper prior knowledge is available (such as the label relation
graph in~\cite{Deng2014}).

To incorporate structured instance-instance correlations, we make one
additional assumption:
\begin{assumption}
\label{as:2} Semantically similar images should have similar
labels.
\end{assumption}
It is reasonable to make this assumption as labels in multi-label image recognition problem can be viewed as a kind of semantic description of images. However, due to the limited label set size and missing labels problem, the observed labels are generally not precise enough. We will discussed this problem in detail in Section~\ref{semantic}.

Assuming that we are able to accurately to project images to the semantic space, we can then incorporate structured instance-instance correlations based on Assumption~\ref{as:2}. Specifically, an undirected weighted graph $G_s=(V_s,E_s,W_s)$ can be constructed with vertices $V_s = \{1,\dots,n\}$ (each vertex corresponds to the semantic representation of one image instance), edges $E_s \subseteq V_s \times V_s$, and the $n\times n$ edge weight matrix
$W_s$ that describes the similarity among image instances in
semantic space. According to Assumption~\ref{as:2}, the learned
label matrix $XM$ on the semantic graph $G_s$ should be smooth. To
be specific, for any two instances $x_i, x_j \in X$, if they are
semantically similar, i.e. the weight $w^s_{i,j}$ of edge
$e^s_{i,j}$ on the semantic graph is large, their labels should
also be similar, i.e., the distance between the learned labels of these two instances should be small. Thus, we define another regularization, aiming to
minimize the distance between the learned labels of any two
semantically similar instances:
\begin{equation}
\label{smooth} \sum_{i,j}w^s_{i,j}\|(x_i-x_j)M\|_2^2 ,
\end{equation}
where $w^s_{i,j}$ is the $\{i,j\}$-th entry of the weight matrix
$W_s$.

\eqref{smooth} is equivalent to
\begin{equation}
\left\|M\right\|_{L_s} \triangleq \label{laplacian} \text{tr}(M^TX^TL_sXM),
\end{equation}
where $L_s = D_s - W_s$ is the Laplacian of graph $G_s$ and $D_s =
\text{Diag}(\sum^n_{j=1}w^s_{ij})$. \eqref{laplacian} is often
referred as the Laplacian regularization term~\cite{Ando2006}. For simplicity, We use $\left\|\cdot\right\|_{L_s}$ to represent to the Laplacian
regularization on $M$ with respect to $L_s$. We
add this regularization term to the multi-label learning
formulation to incorporate structured instance-instance correlations
to the problem. In this way, the objective function of our
multi-label learning with structured instance-instance correlations
becomes:
\begin{equation}
\label{final-obj}
\begin{matrix}
\underset{M}{\min} & F(M)=\lambda\|M\|_{*} + \gamma_s \left
\|M\right\|_{L_s} + \frac{1}{2}
\|\mathcal{R}_{\Omega}(XM)-\tilde{Y}\|_F^2,
\end{matrix}
\end{equation}
where $\gamma_s$ is the trade-off parameter.

If proper structured label-label correlations are available, we
can also incorporate the information by adding another Laplacian
regularization term on $M$ with the label correlation graph.
Specifically, assuming we have an undirected graph
$G_t=(V_t,E_t,W_t)$ with the $c\times c$ weight matrix $W_t$ that
captures the structured label-label correlations, we can similarly
define the corresponding Laplacian regularization as
\begin{equation}
\label{laplacian_label} \left\|M\right\|_{L_t} \triangleq
\text{tr}(XML_tM^TX^T) ,
\end{equation}
where $L_t$ is the Laplacian of the label correlation graph.
However, unlike the label relation graph used in~\cite{Deng2014}
for multi-class classification, the label correlations for
multi-label learning are much more complicated and currently there
is no such information available for multi-label learning, to the
best of our knowledge. Therefore, in this paper, we stick to
\eqref{final-obj} as our optimization objective function. 

The formulation of Zhu et al.~\cite{Zhu2010} is closely related to ours, but with two key differences. Firstly, they focus on solving the tag refinement problem rather than classification. More importantly, our graph construction process is based on relationships in the semantic space with the proposed semantic descriptor rather than in the feature space, which we will describe in the following sections.

\section{Semantic Descriptor Extraction}
\label{semantic} 
As we have discussed, if we are able to represent
the image set with a semantic graph $G_s$, we can incorporate
structured instance-instance correlations to the multi-label learning
problem. The problem now is: how to effectively project the images
to the semantic space and build an appropriate semantic
correlation graph. 

For a multi-label learning problem, the labels of images can be
viewed as semantic descriptions. However, since the size of the
label set for many real-world applications is limited and more
importantly the observed labels could be largely incomplete, using
just the available labels as semantic descriptors would not be
sufficient.

Previous works~\cite{Bergamo2014,Dixit2015,Lampert2014} make use
of the posterior probabilities of the classifications on some general
large-scale datasets such as \textsc{ILSVRC 2012}~\cite{ILSVRC15}
and \textsc{Place} database~\cite{Zhou2014} with large number of
classes as the semantic descriptors. In this paper, we also adopt
such approach and utilize the score vector from CNN trained on
\textsc{ILSVRC 2012} as our global semantic descriptor. To better
adapt the global descriptor to the target domain, we further
develop feature selection to select most relevant semantic
concepts. Moreover, we also propose to pool labels from visual
neighbors of each instance in the target domain as the local semantic descriptor. The
resulting overall semantic descriptor is empirically shown to have
better discriminative power and stability over its individual
components. In the following, we describe the details of the
developed global and local semantic descriptors.

\subsection{Global Semantic Descriptor}
\label{global} 
\begin{figure}
    \centering
    {\includegraphics[width=0.3\textwidth]{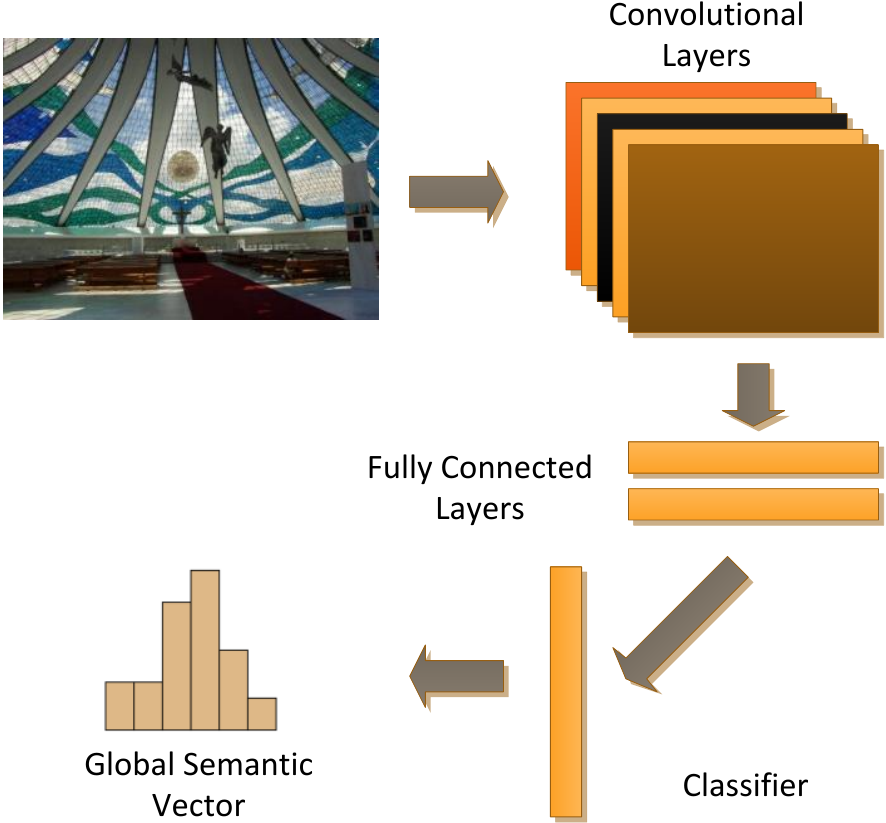}}
\caption{The extraction of the global semantic descriptor using
CNN trained with \textsc{ILSVRC 2012}. Each image is projected to
the semantic space through the convolutional and fully connected
layers of CNN.}
    \label{example-global}
\end{figure}

Given a vocabulary $\mathcal{D} =
\left\{d_1,\dots,d_s\right\}$ of $s$ semantic concepts, a semantic
descriptor of image $x_i$ can be seen as the combination of these
concepts, denoted as $g_i \in \mathbb{R}^s$, $g_i{(j)} \in
\{0,1\}$. As the precise concept combination is not available,
naturally we exploit the score vector extracted from the
classifiers to describe the semantics of an image. Considering
such semantic descriptor is essentially posterior class
probabilities of a given image, we call it \emph{global} semantic
descriptor. Specifically, similar to~\cite{Dixit2015}, we apply
CNN trained with \textsc{ILSVRC 2012} and use the resulting
posterior class probabilities as the global semantic vector. The
process is illustrate in Fig.~\ref{example-global}.
The problem with such global semantic vectors is that many
semantic concepts in the source dataset might not be relevant to
the target dataset. For example, if images from the target dataset
are mainly related to animals, the responses of these images on
some concepts such as man-made objects are generally not helpful
and could even cause confusions. To eliminate such irrelevant or
noisy concepts, we propose a simple feature selection method.
Specifically, let's denote the global semantic descriptions of a
set of $n$ images with respect to concepts $\mathcal{D}$ as
$\tilde{\mathcal{D}} = \left\{\tilde{d}_i \in \mathbb{R}^{n}, i =
1\dots,s\right\}$, and their observed labels $\tilde{Y} =
\left\{\tilde{y}_i^c \in \mathbb{R}^{n}, i = 1\dots,c\right\}$. We
measure the relevance between semantic concept $i$ and the given
label set as:
\begin{equation}
\label{relevance-metric} R_i =
\sum_{j=1}^{c}I(\tilde{d}_i,\tilde{y}_j^c),
\end{equation}
where $I(a,b)$ evaluates the mutual information between $a$ and
$b$. $R_i$ essentially measures the accumulated linear dependency
between concept $i$ and the given labels. After obtaining $R_i$
for all concepts, $\tilde{s}$ concepts are selected based on
descending order of $R_i$ to preserve the most relevant
$\tilde{s}$ concepts for the target dataset. The resulting global
semantic descriptors for the target dataset is then denoted as
$\mathcal{G} = \left\{g_i \in \mathbb{R}^{\tilde{s}}, i =
1\dots,n\right\}$.
\begin{figure*}
    \centering
    {\includegraphics[width=0.85\textwidth]{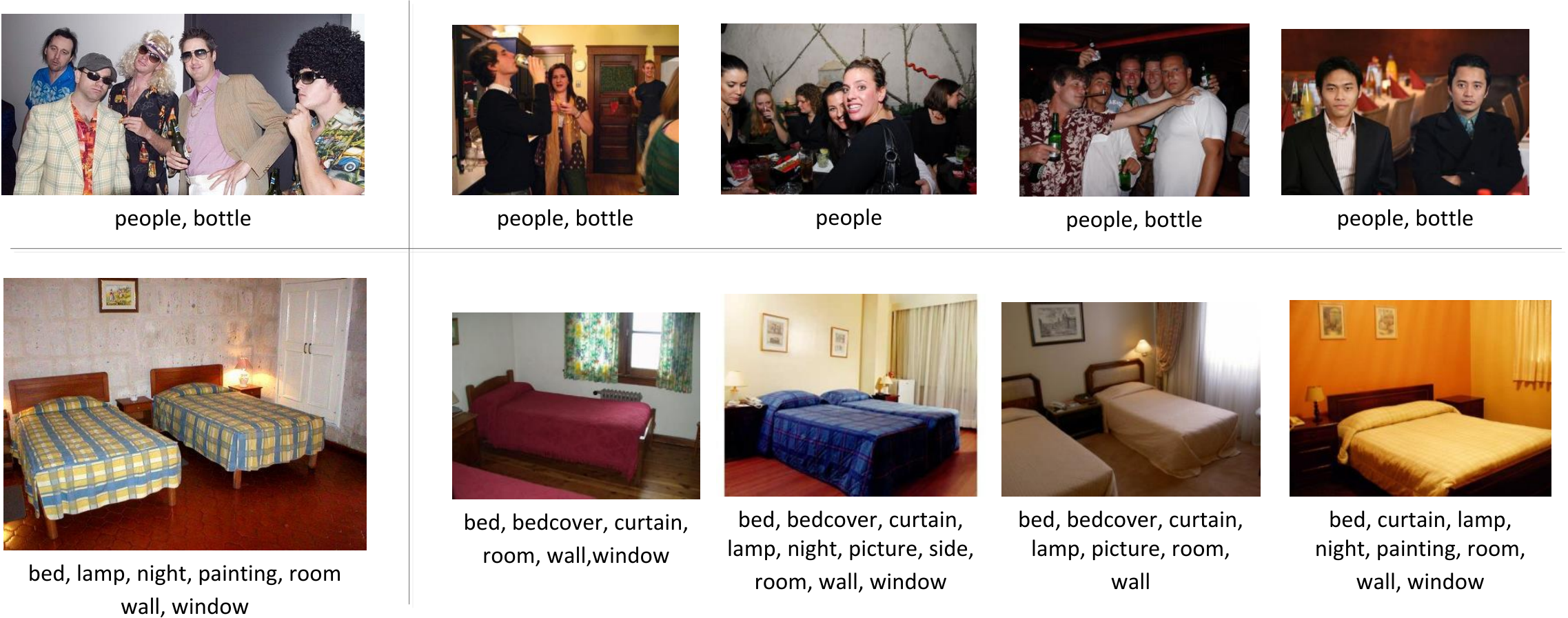}}
\caption{Examples of label relevance between visual neighbors. The
images on the right are the top-$4$ visual neighbors of the images
on the left. The upper images are from \textsc{VOC 2007} and the
bottom images are from \textsc{IAPRTC-12}. As shown here, visual
neighbors are likely to share similar labels.}
    \label{example-local}
\end{figure*}

\subsection{Local Semantic Descriptor}
\label{local} In addition to global semantic descriptor, we
propose to extract local semantic descriptor to enhance the
stability of the semantic descriptor and its relevance to target
labels. Motivated by kNN classification, our basic idea is to
utilize visual neighbors to generate local semantic descriptor. As
illustrated in Fig.~\ref{example-local}, the visual neighbors of
an image are likely to share similar labels. If some labels of a
particular image are missing, it is reasonable to assume that
the observed labels of its visual neighbors can be helpful to
approximate the semantic description of the image. Therefore, we
include labels of visual neighbors as part of our proposed
semantic descriptor.

To be specific, for an image $x_i$, we search for its top-$k_v$
visual neighbors, which have observed labels $\tilde{y}_j^r \in \mathbb{R}^c,
j=1,\dots,k_v$. The local semantic descriptor of $x_i$ is defined
as
\begin{equation}
\label{local-semantic} l_i = \frac{1}{k_v}\sum_{j=1}^{k_v}\tilde{y}_j^r .
\end{equation}
\eqref{local-semantic} is essentially an average pooling of labels
$y_j$, which tells ``what does the image look like''. By find
$l_i$ for all images, we can form a set of local semantic
descriptors $\mathcal{L} = \left\{l_i \in \mathbb{R}^{c}, i =
1\dots,n\right\}$ for the target dataset. The final semantic
descriptor set $\mathcal{S}$ is the direct concatenation of
$\mathcal{G}$ and $\mathcal{L}$, denoted as $\mathcal{S} =
\left\{s_i \in \mathbb{R}^{\tilde{s}+c}, i = 1\dots,n\right\}$ and $s_i^T = \left[g_i^T, l_i^T\right]$.

Note that in order to find accurate visual neighbors, we extract a
low dimensional CNN feature from each image for distance
measurements (see Section~\ref{feature-extraction} for details).
We discuss the effectiveness of the proposed semantic descriptors empirically 
in Section~\ref{validation}.

\subsection{Graph Construction}
\label{graph} After extracting the semantic descriptor set
$\mathcal{S}$, we can now construct the semantic correlation graph
based on $\mathcal{S}$. In particular, we treat each semantic
representation $s_i$ as a node $v_i^s$ of the undirected graph
$G_s$ in the semantic space. To effectively construct the edges
$e_{i,j}^s$ between node $v_i$ and other nodes, following the
general idea of~\cite{Belkin2006}, we first search for $k_s$ neighbors in the semantic space of $v_i$, which we refer as semantic neighbors. Note that the number of semantic
neighbors $k_s$ can be different from the number of visual
neighbors $k_v$ that we use for building local semantic
descriptors. We then connect $v_i$ and its $k_s$ semantic
neighbors to form the edges from $v_i$. The weight of an edge is
defined as the dot-product between its two nodes, i.e.,
\begin{equation}
\label{graph-weight} w^s_{i,j} = s_i^Ts_j.
\end{equation}
The complete process for constructing the semantic correlation
graph is summarized in Algorithm~\ref{graph-alg}.

\begin{algorithm}[t]
    \caption{Semantic correction graph construction}
    \label{graph-alg}
    \begin{algorithmic}[1]
        \STATE \textbf{Input}: A set of images $X = \left\{x_1,\dots,x_n\right\}$ and their corresponding observed labels $\tilde{Y} = \left\{\tilde{y}_i \in \mathbb{R}^{n}, i = 1\dots,c\right\}$.
        \STATE Extract low-dimensional features of $X$ by CNN as  $X^l = \left[x^l_1;\dots;x^l_n\right]^T$, $x^l \in \mathbb{R}^{d_l}$.
        \STATE Extract score vectors of $X$ by CNN trained on \textsc{ILSVRC 2012} as  $X^s = \left[x^s_1;\dots;x^s_n\right]^T$, $x^s \in \mathbb{R}^{s}$. Let $\tilde{\mathcal{D}} = X^s$ and $\tilde{d}_i$ is the $i$-th column vector of $X^s$
        \STATE Calculate $R_i$ in~\eqref{relevance-metric} for all concepts and select top $\tilde{s}$ concepts. The global semantic descriptor is then $\mathcal{G} = \left\{g_i \in \mathbb{R}^{\tilde{s}}, i = 1\dots,n\right\}$, $g_i = x^{\tilde{s}}_i$.
        \STATE Search for top-$k_v$ visual neighbors of each image with low dimensional feature $x^l_i$. Calculate $l_i$ in~\eqref{local-semantic} for all images. The local semantic descriptor is then $\mathcal{L} = \left\{l_i \in \mathbb{R}^{c}, i = 1\dots,n\right\}$.
        \STATE Concatenate $\mathcal{G}$ and $\mathcal{L}$ for $\mathcal{S}$, where $s_i^T = \left[g_i^T, l_i^T\right]$.
        \STATE Search for top-$k_s$ semantic neighbors of each image $s_i$. Connect neighbor nodes as edges. Calculate the weight $w^s_{i,j}$ of edge $e_{i,j}^s$ using \eqref{graph-weight}.
    \end{algorithmic}
\end{algorithm}

\section{Proximal Gradient Descent Based Solver}
Solving our objective function \eqref{final-obj} is not straightforward, although it is convex, the
nuclear norm $\left\|\cdot\right\|_{*}$ is non-smooth.
Following~\cite{Ji2009,Xu2013}, we employ an \emph{accelerated
proximal gradient} (APG) method to solve the problem.

We first consider minimizing the smooth loss function without the
nuclear norm regularization:
\begin{equation}
\label{smooth-obj}
\begin{matrix}
\underset{M}{\min} & f(M) = \gamma_s\left\|M\right\|_{L_s} +
\frac{1}{2} \|\mathcal{R}_{\Omega}(XM)-\tilde{Y}\|_F^2,
\end{matrix}
\end{equation}
A well-known fact~\cite{Bertsekas1999} is that the gradient step
\begin{equation}
\label{grad-step} M_k = M_{k-1} - \mu_k\bigtriangledown f(M_{k-1})
\end{equation}
for solving the smooth problem can be formulated as a proximal
regularization of the linearized function $f(M)$ at $M_{k-1}$ as
\begin{equation}
\label{proximal-step} M_k = \text{arg}\min_M P_{\mu_k}(M,M_{k-1})
\end{equation}
where
\begin{align*}
    P_{\mu_k}(M,M_{k-1})  = f&(M_{k-1})+\left<M-M_{k-1},\bigtriangledown f(M_{k-1})\right> \\
    &+ \frac{1}{2\mu_k}\left\|M-M_{k-1}\right\|_F^2 ,
\end{align*}
$\left<A,B\right>= tr(A^TB)$ denotes the matrix inner product, and
$\mu_k$ is the step size of iteration $k$.

Based on the above derivation, following~\cite{Ji2009},
\eqref{final-obj} is then solved by the following iterative
optimization:
\begin{equation}
    \label{iter-step}
    M_k  = \text{arg}\min_M Q_{\mu_k}(M,M_{k-1})\triangleq P_{\mu_k}(M,M_{k-1}) + \lambda\left\|M\right\|_{*}.
\end{equation}
Further ignoring the terms that do not dependent on $M$, we
simplify~\eqref{iter-step} into minimizing
\begin{equation}
    \label{final-iter-step}
    \frac{1}{2\mu_k}\left\|M-\left(M_{k-1}-\mu_k\bigtriangledown
    f(M_{k-1})\right)\right\|_F^2+\lambda\left\|M\right\|_{*},
\end{equation}
which can be solved by singular value thresholding (SVT)
techniques~\cite{Cai2008}.

Algorithm~\ref{apggraph} shows the APG method we used for
solving~\eqref{final-obj}. Similar to~\cite{Xu2013}, we introduce
an auxiliary variable $V$ (line $4$) to accelerate the
convergence. At each step, by utilizing the Lipschitz continuity
of the gradient of $f(\cdot)$, the step size $\mu_k$ can be found
in an iterative fashion. Specifically, we start from a constant
$\mu_1 = A$ and iteratively increase $\mu_k$ until the following
condition is met:
\begin{equation}
\label{stepsize_condition} F(M_k) \leq Q_{\mu_k}(M_k,M_{k-1})
\end{equation}
which is equivalent to line 6 in Algorithm~\ref{apggraph}.

\begin{algorithm}[t]
    \caption{APG-Graph}
    \label{apggraph}
    \begin{algorithmic}[1]
        \STATE \textbf{Initialization}: $\theta_1 = \theta_2 \in (0,1]$, $M_1 = M_2$, $\mu = A$, $\rho>1$, and stopping criterion $\epsilon$
        \STATE $k=2$;
        \WHILE{$F(M_{k+1})\leq (1-\epsilon)F(M_{k})$}
            \STATE $V_k = M_k + \theta_k(\theta_{k-1}^{-1}-1)(M_k-M_{k-1})$
            \STATE $M_{k+1} = \text{arg}\min_M Q_{\mu}(M,V_{k})$
            \WHILE{$F(M_{k+1})> Q_{\mu}(M_{k+1},V_k) $}
                \STATE $\mu=\mu*\rho$
                \STATE $M_{k+1} = \text{arg}\min_M Q_{\mu}(M,V_k)$
            \ENDWHILE

            \STATE $\theta_{k+1}=(\sqrt{\theta_k^4+4\theta_k^2}-\theta_k^2)/2$
            \STATE $k=k+1$
        \ENDWHILE
    \end{algorithmic}
\end{algorithm}

\section{Experimental Results}
\label{exp-sec} In this section, we compare our proposed APG-Graph
algorithm with several state-of-the-art methods on four widely
used multi-label learning benchmark datasets.  The details of the
benchmark datasets can be found in Table~\ref{datasets}. We follow
the pre-defined split of \textsc{train} and
\textsc{test}\footnote{
\url{http://lear.inrialpes.fr/people/guillaumin/data.php}}. To
mimic the effect of missing labels, we uniformly sample $\omega\%$
of labels from each class of the \textsc{train} set, where $\omega
\in \{10,20,30,40,50\}$. It means we only use $10-50\%$ of the
training labels. We use mean average precision (mAP) as our
evaluation metric, which is the mean of average precision across
all labels/tags of the \textsc{test} set and is widely used in
multi-label learning.

NUS-WIDE is also widely used as multi-label classification benchmark dataset. Unfortunately, we cannot obtain all the images from NUS-WIDE dataset. Since we are unable to extract the semantic descriptors without original images, we cannot perform experiments in this dataset.

\begin{table}[h]
 \centering
 \caption{Dataset Information} \label{datasets}
 \begin{tabular}{c | r r r r }
  \hline
  Dataset &\#Train &\#Test &\#Labels &\#Avg Labels \\ \hline
  \textsc{VOC 2007} &5011 &4952 &20 &1.4\\
  \textsc{ESP Game} &18689 &2081 &268 &4.5 \\
  \textsc{Flickr 25K} &12500 &12500 &38 &4.7 \\
  \textsc{IAPRTC-12} &17665 &1962 &291 &5.7 \\
 \end{tabular}
\end{table}

\subsection{Experiment Setup}
\label{feature-extraction}

{\bf{Feature representation for input data $X$}:} For all the
image instances (\textsc{train} and \textsc{test}), we need to
find their effective feature representations as the input data
$X$. Note that for simplicity, we abuse the notation $X$ for both
the input image set and the corresponding image description set.
In particular, we employ the 16-layer very deep CNN model
in~\cite{Simonyan2014}.  We apply the CNN pre-trained on
\textsc{ILSVRC 2012} dataset to each image and use the activations
of the $16$-th layer as the visual descriptor ($4096$-dimensional)
of the image. We then concatenate the semantic descriptor $s_i$
developed in Algorithm~\ref{graph-alg} with this
$4096$-dimensional visual descriptor as the overall feature
representation for image $x_i$. To satisfy our
Assumption~\ref{as:2}, we further apply PCA to the overall feature
representations to decorrelate the features. The dimension of PCA
features is set to preserve $90\%$ energy of the original
features, which results in the final descriptor of dimensions
around $700$.

{\bf{Finding visual neighbors:}} To find accurate visual neighbors
for local semantic descriptor, we extract a low-dimensional CNN
descriptor for each image. We use the same 16-layer very deep CNN
structure, except that the activations of the $16$-th fully
connected layer is of $128$ dimensions instead of $4096$. The
$128$-d descriptors denoted as $X^l$ are used to find visual neighbors as described
in Section~\ref{local}.

{\bf{Baselines:}} We compare our method with the following
baselines. \squishlist \item \textsc{Maxide}~\cite{Xu2013}: A
matrix completion based multi-label learning method using training
data as side information to speed up the training process.
Although the formulation of \textsc{Maxide} incorporate a label
correlation matrix, while in experiments \textsc{Maxide} actually
sets it as identity matrix. \textsc{Maxide} outperforms other matrix
completion based methods like MC-$1$ and
MC-b~\cite{Kong2013,Cabral2011}. The formulation of \textsc{Maxide} is similar to our formulation without the Laplacian regularization term.

\item \textsc{FastTag}~\cite{Chen2013}: A fast image tagging
algorithm based on the assumption of uniformly corrupted labels.
\textsc{FastTag} learns an extra transformation on the label
matrix to recover its missing entries. It achieves
state-of-the-art performances on several benchmark datasets.

 \item \textsc{Binary Relevance}~\cite{Tsoumakas2007}: \textsc{BR} is a popular baseline for multi-label classification. It treats each class as a separate binary classification to solve the problem. Here we consider linear binary relevance and use \textsc{LIBLINEAR}~\cite{Fan2008} to train a binary classifier for each class.

 \item \textsc{Least Squares}: \textsc{LS} is a a ridge regression model which uses the partial subset of labels to learn the decision parameter $M$.

\squishend We cross-validate the parameters of these methods on
smaller subsets of benchmark datasets to ensure best performance.

{\bf{Our parameters:}} The learning part of our method has two
parameters $\gamma_s$ and $\lambda$ as shown in~\eqref{final-obj}.
Similar to other methods, we cross-validate on a small subset of
benchmark datasets to get the best parameters. The parameters for
the semantic correlation graph construction are decided
empirically. Specifically, the number of semantic concepts
$\tilde{s}$ used in global semantic descriptors is set to be $0.5\
c$. The number of visual neighbors $k_v$ is set to be $50$ and the
number of semantic neighbors $k_s$ is set to be $10$.
\begin{figure}
\centering
  \subfloat[mAP]{\includegraphics[width=0.35\textwidth]{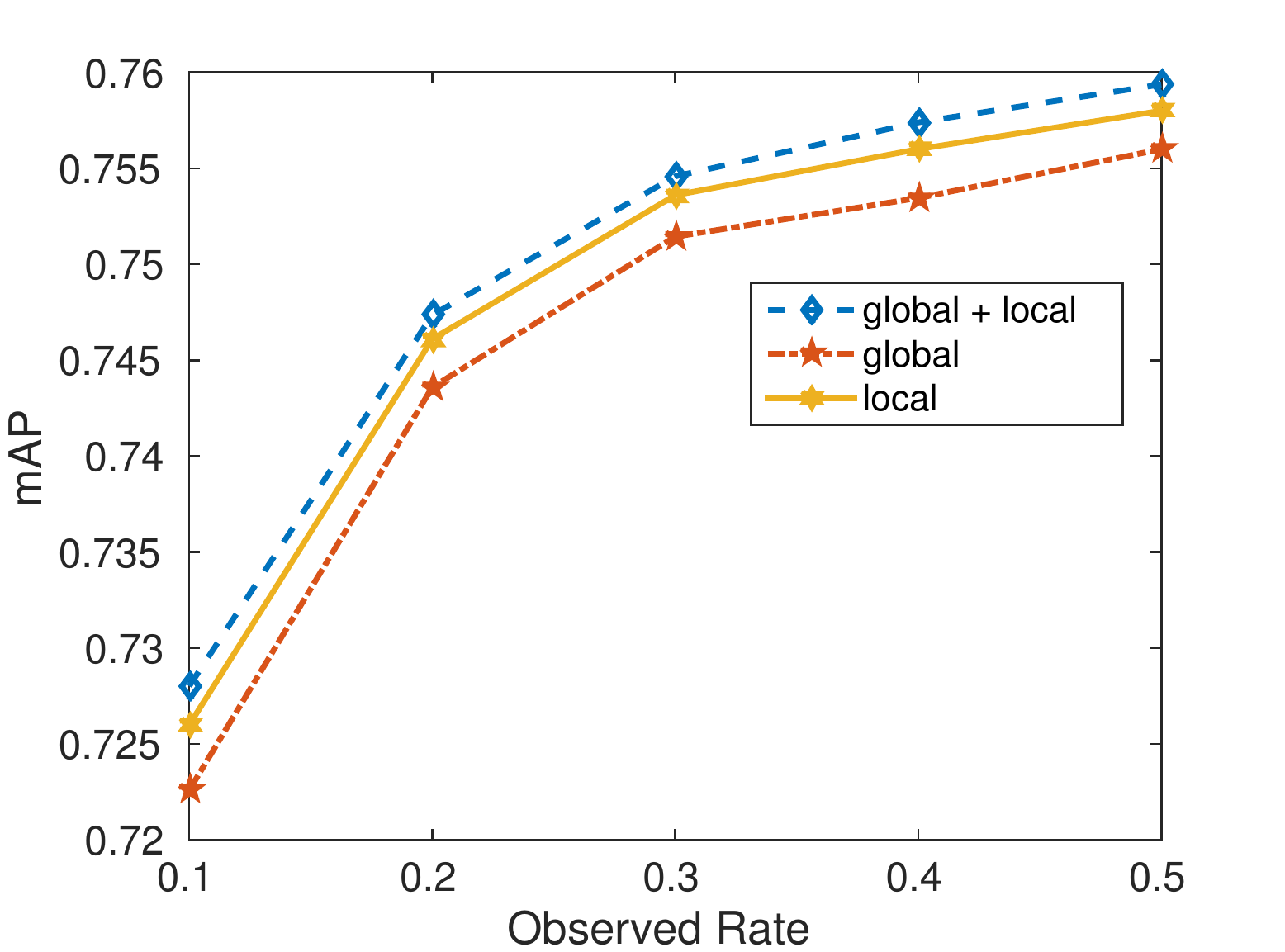}}
  \caption{Validation experiments of the three semantic descriptors on Flickr25K dataset. We can see from the mAP that the proposed global + local semantic descriptor achieves the best performance}
  \label{val-exp}
\end{figure}
\subsection{Validation of Semantic Descriptor}
\label{validation}

We validate the effectiveness of the proposed semantic descriptor on Flickr25K dataset by demonstrating the classification accuracy.  As shown in Fig.~\ref{val-exp}, for the recognition rate on the test set, our proposed global + local descriptor has the highest mAP consistently. The gain over just using local semantic descriptor is not so large though. We suspect that since the global semantic descriptors are extracted from \textsc{ILSVRC} dataset, which is an object dataset, and the tags of Flickr25K are mostly \emph{not} related to objects, the global semantic descriptor is not so helpful in this case. If we use other sources of global semantic vocabulary more related to scene, e.g., \textsc{Place} database, we could potentially have even better performance.

\subsection{Comparison with Other Methods}

\begin{figure*}
\centering
  \subfloat[\textsc{VOC 2007}]{\includegraphics[width=0.32\textwidth]{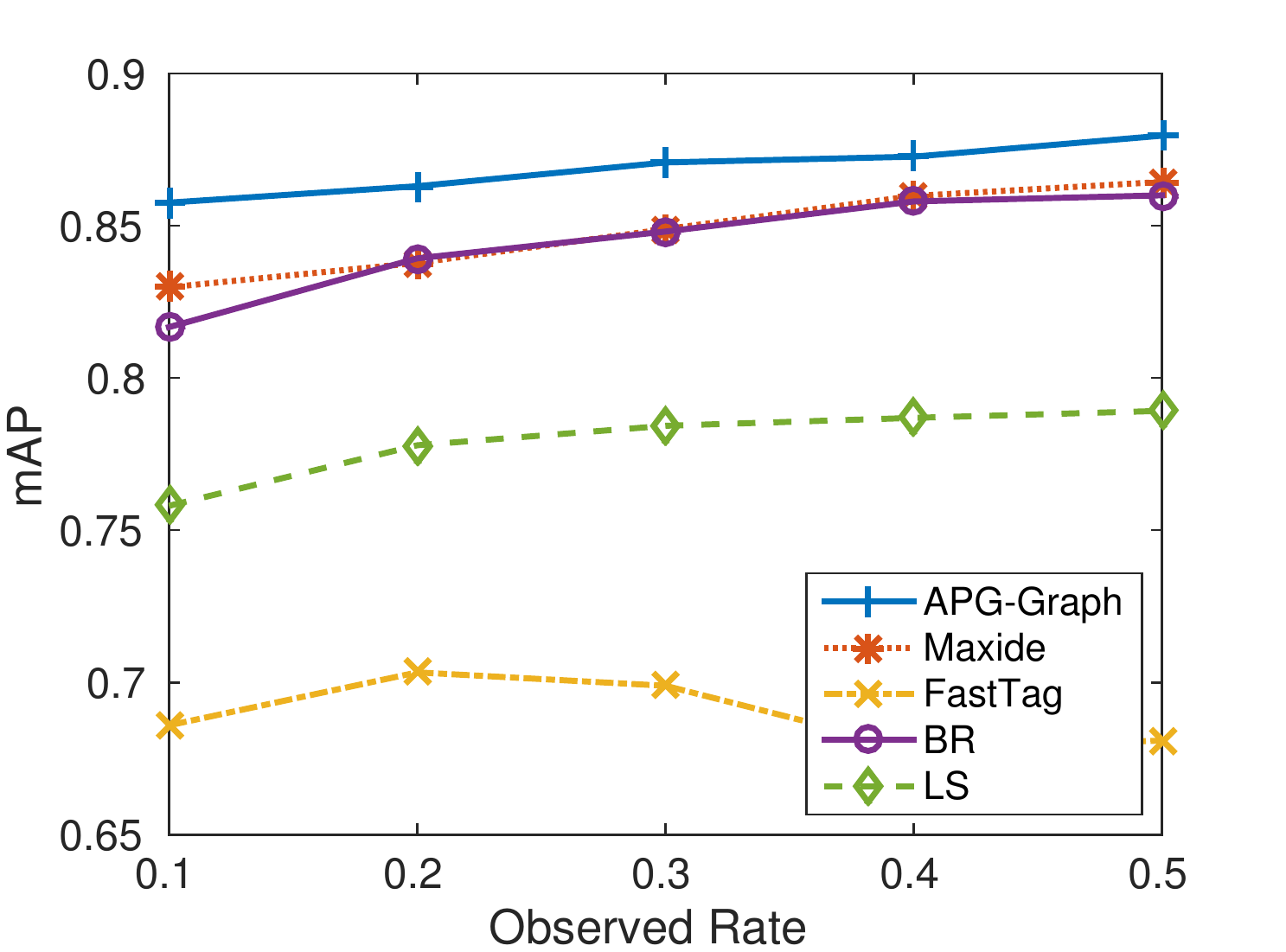}}
  \subfloat[\textsc{Flickr25K}]{\includegraphics[width=0.32\textwidth]{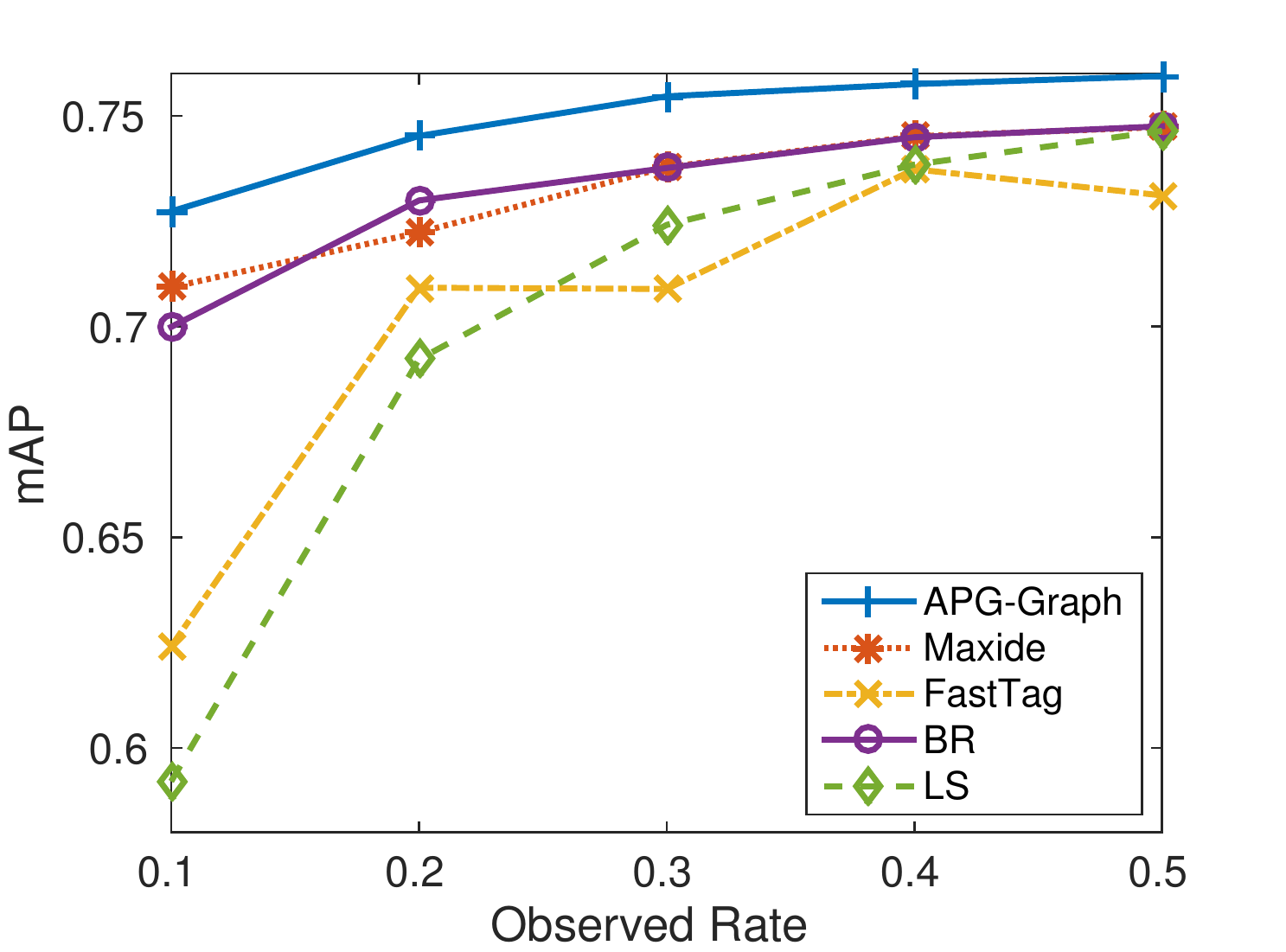}}\\
  \subfloat[\textsc{ESP Game}]{\includegraphics[width=0.32\textwidth]{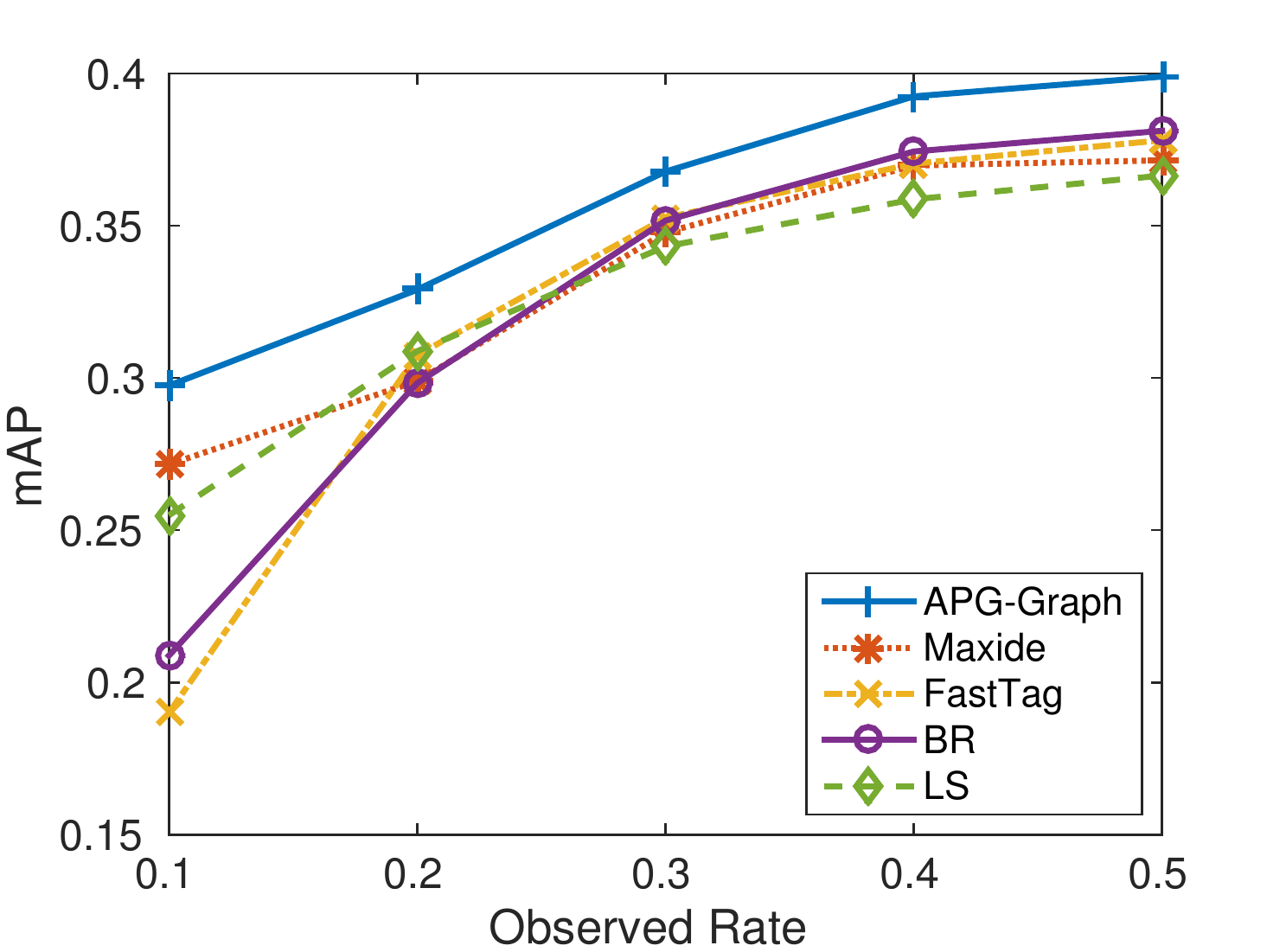}}
  \subfloat[\textsc{IAPRTC-12}]{\includegraphics[width=0.32\textwidth]{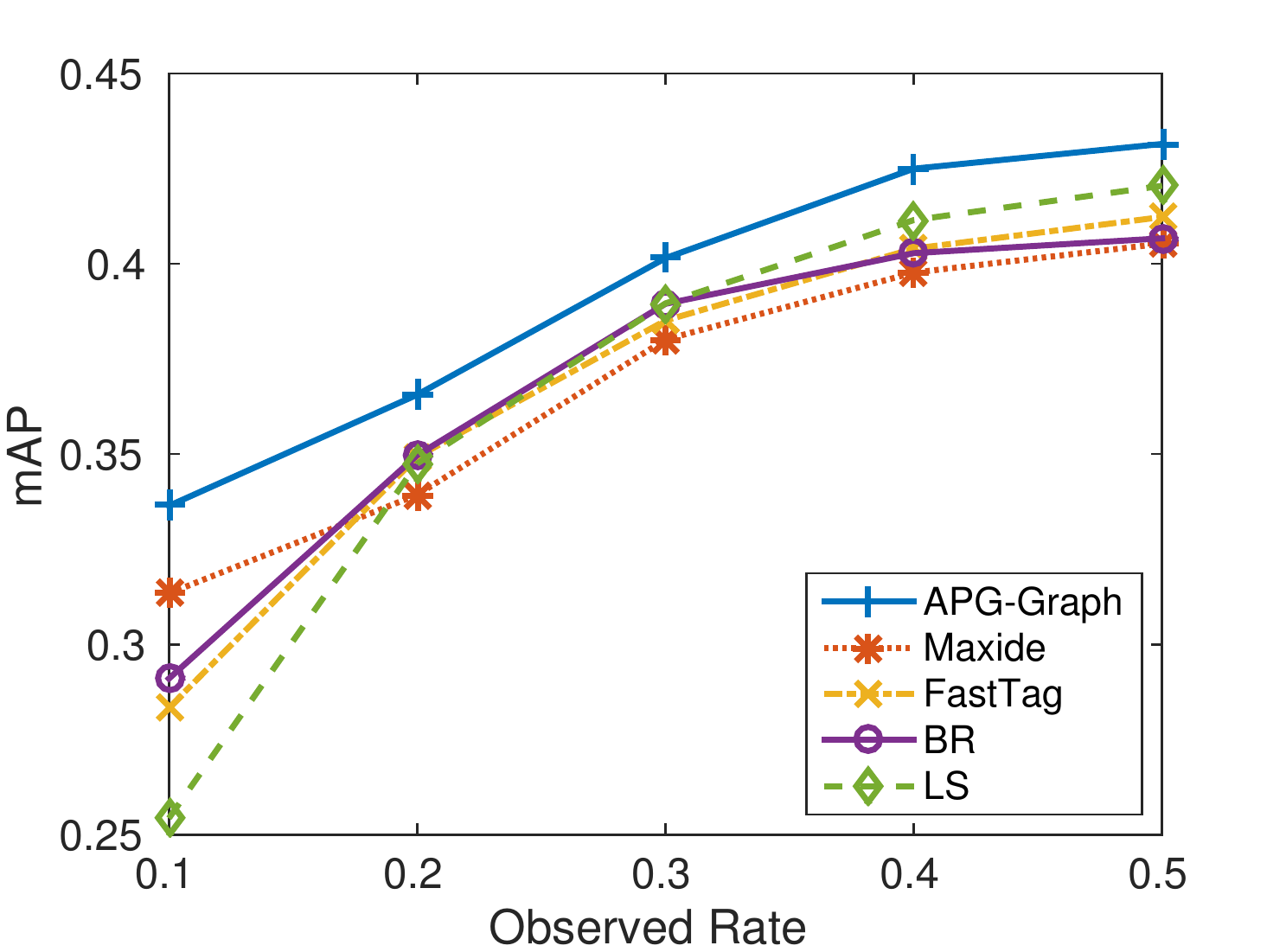}}
  \caption{The mAP Results (in \%) of different methods on the four benchmark datasets with observed label rates ranging from $0.1$ to $0.5$.}
  \label{exp}
\end{figure*}

\begin{figure*}
    \centering
    {\includegraphics[width=0.75\textwidth]{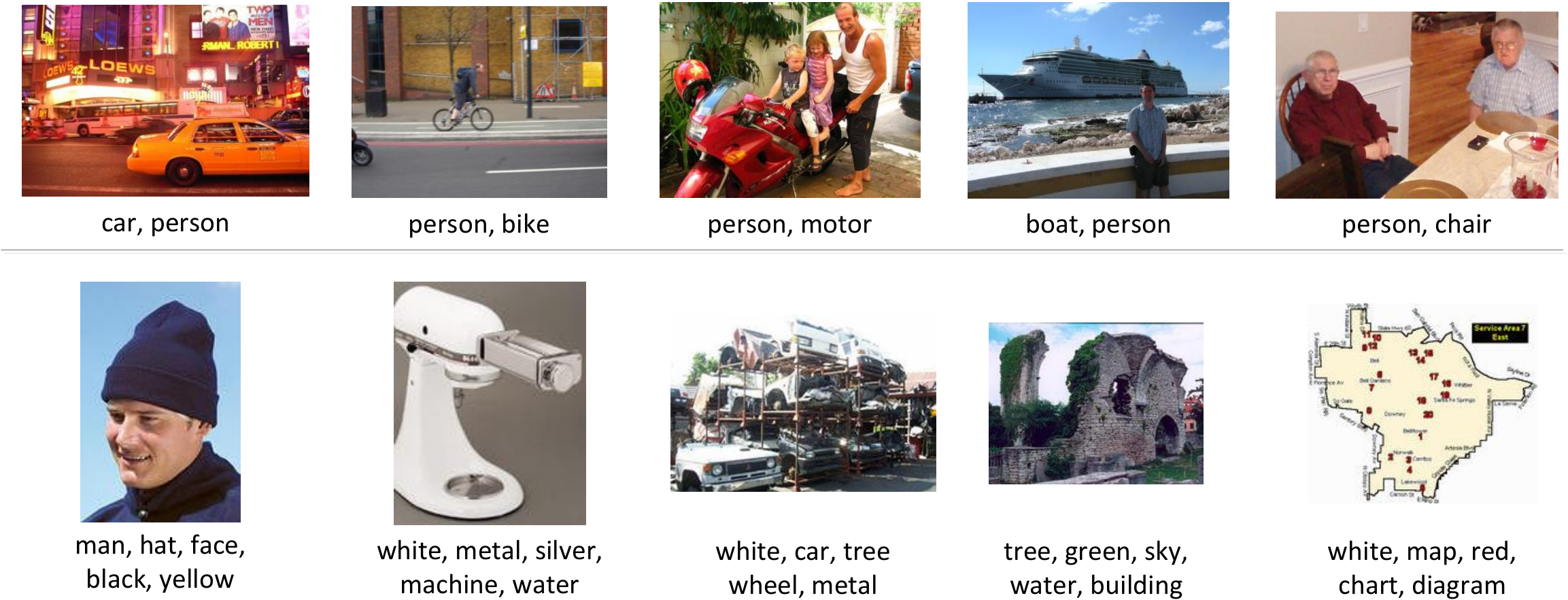}}
    \caption{Examples of generated labels using our proposed APG-Graph method. We only observe $10\%$ of the given labels in the training set. The upper images are randomly selected from the test set of \textsc{VOC 2007} with top-$2$ labels shown. The bottom images are randomly selected from the test set of \textsc{ESP Game} with top-$5$ labels shown. As we can see, the labels accurately match the images.}
    \label{example-recognition}
\end{figure*}

Fig.~\ref{exp} shows the mAP results of our proposed method and the four baselines on the four benchmark datasets. It can be seen that our method
(\textsc{APG-Graph}) constantly outperforms other methods,
especially when the observed rate is small. The performance gain
validates the effectiveness of our proposed semantic descriptors
and the usage of structured instance-instance correlation. On the other
hand, \textsc{Maxide} generally achieves similar recognition rate
as \textsc{BR} for observed rates ranging from $0.2$ to $0.5$ while it
outperforms \textsc{BR} at an observed rate of $0.1$, which
suggests that the unstructured correlation enforced by the
low-rank constraint (nuclear norm) is helpful at small observed
rates, but the effect is similar to the L$2$ norm used
in SVM classification at large observed label rates. We
use the code provided by~\cite{Chen2013} for $\textsc{FastTag}$.
It seems that \textsc{FastTag} is not very effective in our
experiments, especially for datasets with fewer labels
(\textsc{VOC2007} and \textsc{Flickr25K}). We suspect that the
hyper-parameter tuning in \textsc{FastTag} is not stable when the
labels are fewer. We also show some examples of recognized images
in Fig.~\ref{example-recognition}.
Note that other methods such as TagProp~\cite{Guillaumin2009} and TagRelevance~\cite{Li2009} are not designed for our problem setting and cannot handle missing labels properly, thus in our preliminary experiments their results are bad and we choose not to report them. 

\section{Conclusion}
In this paper, we have incorporated structured semantic correlations to
solve the missing label problem of multi-label learning.
Specifically, we project images to the semantic space with an
effective semantic descriptor. A semantic graph is then
constructed on these images to capture the structured correlations
between images. We utilize the semantic graph Laplacian as a
smooth term in the multi-label learning formulation to incorporate
these correlations. Experimental results demonstrate the
effectiveness of our proposed multi-label learning framework as well as our proposed semantic representation. Future works could include utilizing other large scale datasets such as
\textsc{Place} as another source of global semantic concepts and incorporating structured label correlations.

\noindent\textbf{Acknowledgments} \small This research is supported by Singapore MoE AcRF Tier-1 Grant RG138/14 and also partially by Rolls-Royce@NTU Corportate Lab Project C-RT3.5. The Tesla K40 used for this re-search was donated by the NVIDIA Corporation.

{\small
\bibliographystyle{splncs}
\bibliography{egbib}
}
\end{document}